\documentclass[10pt,twocolumn,letterpaper]{article}

\usepackage{iccv}
\usepackage{times}
\usepackage{epsfig}
\usepackage{graphicx}
\usepackage{amsmath}
\usepackage{amssymb}

\usepackage{multirow}
\usepackage{makecell}
\usepackage{mwe}
\usepackage{graphbox}

\DeclareMathOperator{\rank}{rank}

\newcommand{\bs}[1]{\mathbf{#1}}

\DeclareMathOperator*{\argmin}{argmin}
\newcommand{\norm}[1]{\left \Vert#1 \right \Vert}
\usepackage[breaklinks=true,bookmarks=false]{hyperref}

\iccvfinalcopy % *** Uncomment this line for the final submission

\def\iccvPaperID{4672} % *** Enter the ICCV Paper ID here

\ificcvfinal\fi

\begin{document}

%%%%%%%%% TITLE
\title{Jointly Aligning Millions of Images with \\Deep Penalised Reconstruction Congealing}

\author{Roberto Annunziata\\
Onfido, UK\\
% London, UK\\
{\tt\small roberto.annunziata@onfido.com}
\and
Christos Sagonas\\
Onfido, UK\\
% London, UK\\
{\tt\small christos.sagonas@onfido.com}
\and
Jacques Cali\thanks{Contribution to this research project was entirely made while this co-author was at Onfido.}\\
Blue Prism\\
% London, UK\\
{\tt\small jqscali@gmail.com}
}

\maketitle
\ificcvfinal\thispagestyle{empty}\fi

%%%%%%%%% ABSTRACT
\begin{abstract}
   Extrapolating fine-grained pixel-level correspondences in a fully unsupervised manner from a large set of misaligned images can benefit several computer vision and graphics problems, e.g. co-segmentation, super-resolution, image edit propagation, structure-from-motion, and 3D reconstruction. Several joint image alignment and congealing techniques have been proposed to tackle this problem, but robustness to initialisation, ability to scale to large datasets, and alignment accuracy seem to hamper their wide applicability. To overcome these limitations, we propose an unsupervised joint alignment method leveraging a densely fused spatial transformer network to estimate the warping parameters for each image and a low-capacity auto-encoder whose reconstruction error is used as an auxiliary measure of joint alignment. Experimental results on digits from multiple versions of MNIST (i.e., original, perturbed, \mbox{affNIST} and \mbox{\mbox{\mbox{infiMNIST}}}) and faces from LFW, show that our approach is capable of aligning millions of images with high accuracy and robustness to different levels and types of perturbation. Moreover, qualitative and quantitative results suggest that the proposed method outperforms state-of-the-art approaches both in terms of alignment quality and robustness to initialisation.   
\end{abstract}

%%%%%%%%% BODY TEXT
\section{Introduction}
Establishing pixel-level correspondences between pair of images including instances of the same object category can benefit several important applications, such as motion estimation~\cite{pan2015efficient}, medical imaging~\cite{annunziata2016fully,iglesias2018joint}, object recognition~\cite{girshick2015fast} and 3D reconstruction~\cite{izadi2011kinectfusion}. As a result, this has become a fundamental problem in computer vision~\cite{forsyth2003modern, hartley2003multiple}. Typically, pixel-level correspondences between two images are computed by extracting sparse local feature descriptors (e.g., SIFT~\cite{lowe2004distinctive}, HOG~\cite{dalal2005histograms}, SURF~\cite{bay2006surf}, SCIRD~\cite{annunziata2015scale, annunziata2016accelerating}), then matching the extracted descriptors, and finally pruning mismatches based on geometric constraints. Although this approach has been applied successfully in different domains, its performance can degrade significantly due to factors such as intra-class variations, non-rigid deformations, partial occlusions, illumination, image blur, and visual clutter. Recently, the representational power of Convolutional Neural Networks (CNNs) has been leveraged to improve the overall process. In particular, several CNN-based methods for learning powerful feature descriptors have been introduced~\cite{simo2015discriminative, zagoruyko2015learning, noh2017largescale}. More recently, end-to-end trainable CNNs for learning image descriptors as well as estimating the geometric transformation between the two images have been introduced in~\cite{rocco2019convolutional, kanazawa2016warpnet}.
\begin{figure}[!t]
\centering
\setlength\tabcolsep{1.5pt}
\begin{tabular}{cc}
\includegraphics[align=c, width=0.225\textwidth]{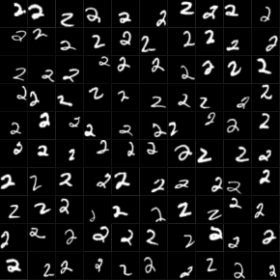} & \includegraphics[align=c, width=0.225\textwidth]{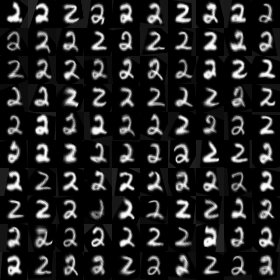}\vspace{-0.3cm}\\\\
\includegraphics[align=c, width=0.225\textwidth]{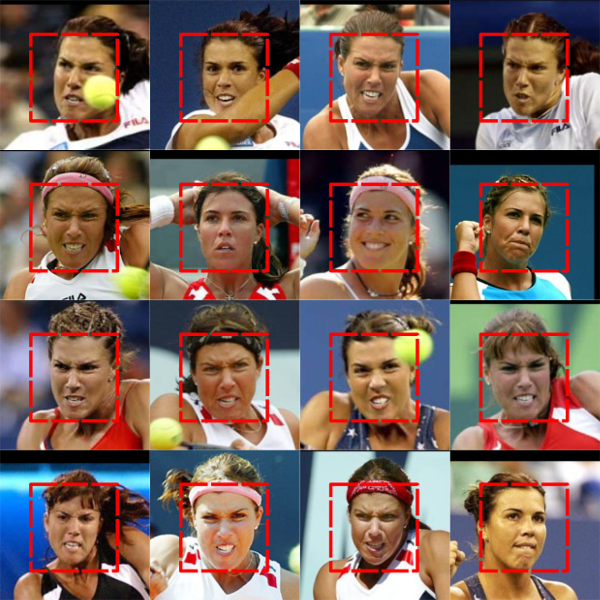} & \includegraphics[align=c, width=0.225\textwidth]{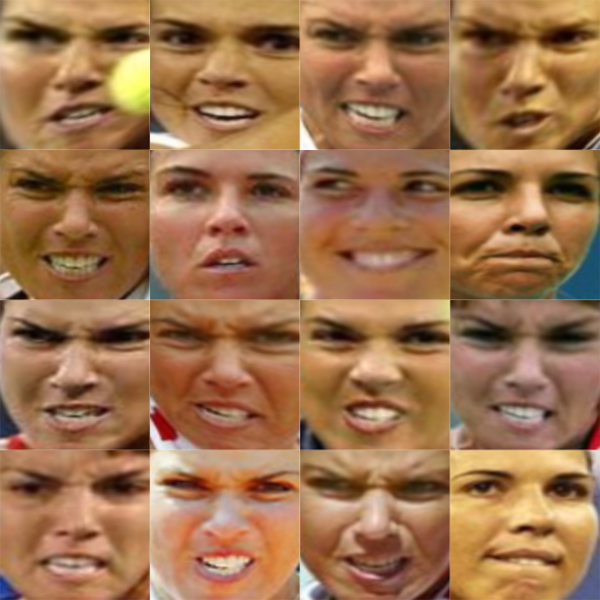} \\
(a) & (b) 
\end{tabular}
\caption{Unsupervised joint alignment (a.k.a. \textit{congealing}) results obtained by the proposed method on digit `$2$' from \textit{affNIST}~\cite{afnist} and \texttt{Jennifer\_Capriati} from LFW~\cite{huang2008labeled}. (a) input images before alignment (initialisation in red), (b) output images aligned with the proposed method.}
\label{fig:main_page}
\end{figure}
The majority of previously proposed methods focus on the problem of finding pixel-level correspondences between a \textit{pair} of images. However, a plethora of other tasks such as, co-segmentation, image edit propagation, video stabilisation and structure-from-motion, require global correspondences between a \textit{set} of images containing a specific object. A straightforward way to address this problem is to identify pixel correspondences between each pair of images in the dataset and solve the problem in a sequential manner. However, this approach would be prone to important limitations, as (i)~it would fail to take into account valuable cross-image appearance information (i.e. statistics of local patches across the entire dataset) during the optimisation; and (ii)~the computational complexity of the problem would increase exponentially with the number of images, therefore significantly limiting the scalability to large datasets. Thus, estimating global correspondences of a set of images (image ensemble) by jointly aligning them in an unsupervised manner can be immensely valuable.

Congealing (joint image alignment) was originally introduced by Learned-Miller in~\cite{learned2006data}. His approach aligns (by estimating rigid transformations) an ensemble of images of a particular object by minimising the sum of entropies of pixel values at each pixel location. Although this method has been effectively applied to handwritten digits and magnetic resonance image volumes, it has shown some limitations, including slow and/or sometimes poor convergence and relatively high sensitivity to hyper-parameters. Later, Huang~\textit{et al.} improved the performance of~\cite{learned2006data} by using hand-crafted SIFT features~\cite{huang2007unsupervised}. To overcome the optimisation problems of the original congealing approach, Cox \textit{et al.}~\cite{cox2008least, cox2009least} proposed to utilise a reference image (i.e., template) and then minimise the sum of squared differences instead of the sum of entropies. This way, standard Gauss-Newton gradient descent method could be adopted to make the optimisation efficient. Later, motivated by lossy compression principles, Vedaldi \textit{et al.}~\cite{vedaldi2008joint} proposed a joint alignment approach based on log-determinant estimation. 

A common drawback of the aforementioned methods is that they cannot simultaneously handle variability in terms of illumination, gross pixel corruptions and/or partial occlusions. RASL, an image congealing method that overcomes this drawback was proposed in~\cite{peng2012rasl}. The key assumptions made in RASL and its multiple variants, e.g.~\cite{likassa2018modified, chen2016nonconvex, oh2015partial} are that (i) an ensemble of well-aligned images of the same object is approximately low-rank and (ii) that gross pixel errors are sparsely distributed. Therefore, image congealing is performed by seeking a set of optimal transformations such that the ensemble of misaligned images is written as the superposition of two components i.e., a low-rank component and a sparse error component. RASL has been widely used for jointly aligning multiple images in different applications such as face landmarks localisation~\cite{sagonas2014raps, peng2015piefa, sagonas2017robust}, pose-invariant face recognition~\cite{sagonas2017robust, sagonas2015robust}, and medical imaging~\cite{bise2016vascular}. Despite its wide applicability, it is worth noting that (i)~RASL joint alignment performance can be severely affected by non-optimal initialisation and high intra-class variability in the image ensemble; (ii)~scalability to large ensembles is limited by the formulation of the low-rank minimisation problem and related SVD-based sub-routines; and (iii)~a new optimisation is required for every new image added to the ensemble. To address some of these limitations, t-GRASTA~\cite{he2014iterative} and PSSV~\cite{oh2015partial} have been recently proposed.

The first deep learning approach to unsupervised joint image alignment was proposed by Huang \textit{et al.}~\cite{huang2012learning}. A modified version of the convolutional restricted Boltzmann machine was introduced to obtain features that could better represent the image at differing resolutions, and that were specifically tuned to the statistics of the data being aligned. They then used those learnt features to optimise the standard entropy-based congealing loss and achieved excellent joint alignment results on the Labelled Faces in the Wild (LFW) benchmark.
\begin{figure}[!tb]
\centering
\includegraphics[align=c, width=0.47\textwidth]{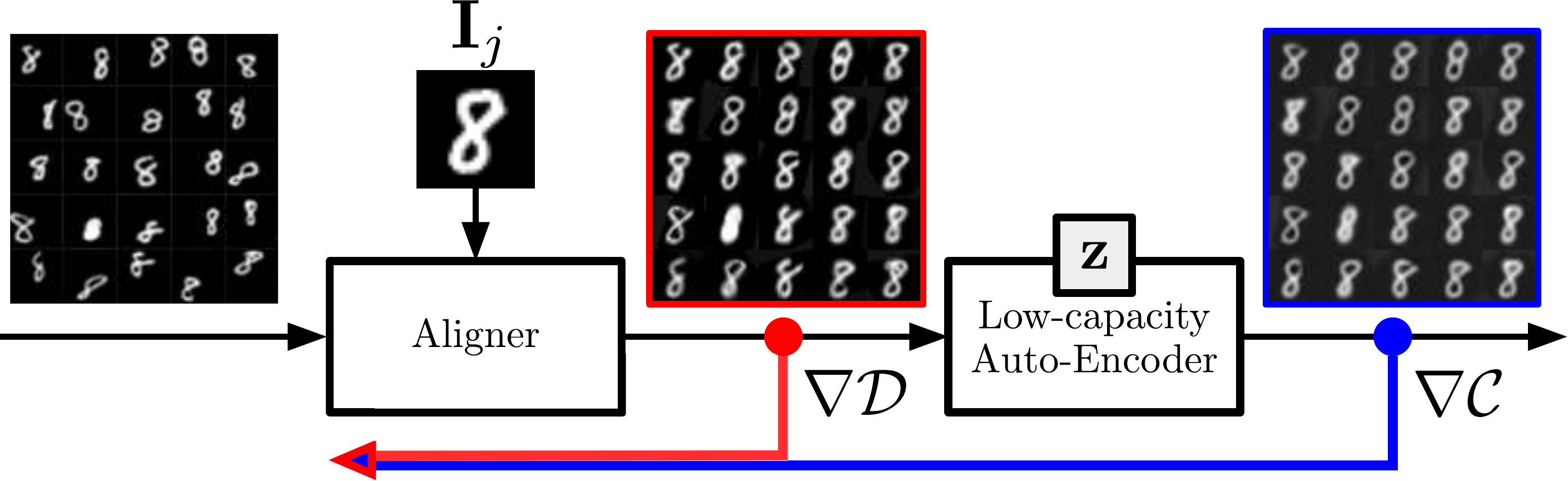}
\caption{Block diagram of the proposed method. Black arrows correspond to forward pass, while red and blue to back-propagation.}
\label{fig:flow_chart}
\end{figure}

Here, we propose a congealing method to solve large-scale joint alignment problems, which is of significant practical importance in light of the ever increasing availability of image data. The proposed method consists of two main modules: (i)~the aligner and (ii)~the low-capacity auto-encoder. Specifically, the joint alignment task is cast as a batch-based optimisation problem in which the aligner is used to estimate the global transformation required to warp each image to a reference. The alignment error is quantified via $\ell_1$-norm between the transformed batch images and the reference. Motivated by the observation that a set of well-aligned images require less modelling capacity to be reconstructed well (e.g. reconstruction with low-rank bases~\cite{peng2012rasl}), the aligned batch is subsequently processed by a \textit{low-capacity} auto-encoder and reconstruction errors are back-propagated to the aligner (a snapshot of the results is displayed in Fig.~\ref{fig:main_page}).

\textbf{Contributions:} In summary, the main contributions of this paper are: (i)~a congealing method which is shown to be capable of handling large-scale joint alignment problems i.e., up to one million data points, simultaneously; (ii)~a novel \textit{differentiable} formulation of the congealing problem, which combines the advantages of previously proposed similarity- and rank-based approaches and that can be easily optimised with Stochastic Gradient Descent (SGD), end-to-end; (iii)~an extensive experimental evaluation of the proposed method and state-of-the-art approaches on several benchmark datasets, including digits and faces at different resolutions, assessing joint alignment performance and robustness to linear and non-linear geometric perturbations of different magnitude and type.
%-------------------------------------------------------------------------
\section{Methodology}
In the following, we briefly summarise the approaches most related to ours. Then, we introduce the proposed method.

\textbf{RASL.}~Let us assume we have $N$ misaligned images $\{\bs{I}_i\}_{i=1}^N \in \mathbb{R}^{w\times h}$ of a particular object and let $\{\bs{p}_i\}_{i=1}^N$ be a set of transformations such that $\{\bs{I}_i^0 = \bs{I}_i\circ\bs{p}_i\}_{i=1}^N$ becomes a set of well-aligned images. If we define ${\text{vec}:\mathbb{R}^{w\times h} \rightarrow \mathbb{R}^m}$ as the operator that vectorises an image, the main assumption of the RASL method is that the matrix:
\begin{equation}
    \bs{D}\circ\bs{P} = [\text{vec}(\bs{I}_1^0)~|~\cdots~|~\text{vec}(\bs{I}_N^0)] = \bs{A}
\end{equation}
will be approximately \textit{low-rank}. However, in practice this assumption can be violated when the object of interest is affected by occlusion, shadows, and noise. Therefore, the authors assume that each aligned image is corrupted with non-Gaussian-but-sparse errors $\bs{E}\in \mathbb{R}^{m\times N}$, such that $ \bs{D}\circ\bs{P} =  \bs{A} + \bs{E}$.
% %
% \begin{equation}
%     \bs{D}\circ\bs{P} =  \bs{A} + \bs{E}.
% \end{equation}
% %
Given the observation of the misaligned and corrupted images, the goal is to estimate a set of transformations $\{\bs{p}_i\}_{i=1}^N$ such that the rank of the transformed noise-free images $\{\bs{I}_i\circ\bs{p}_i\}_{i=1}^N  \in \mathbb{R}^{w\times h}$ becomes as small as possible. Formally,
\begin{equation}\label{eqn:RASL}
    \argmin_{\bs{A},\bs{E},\{\bs{p}_i\}_{i=1}^N} \rank(\bs{A}) \quad \textrm{s.t.}~~\bs{D}\circ\bs{P} = \bs{A} + \bs{E},~\norm{\bs{E}}_0 \leq q,
\end{equation}
where $q$ controls the sparsity of the error matrix $\bs{E}$. 
Unfortunately, the non-convex and discontinuous nature of the optimisation problem in Eq.~(\ref{eqn:RASL}) makes it not directly tractable. To this end, an algorithm that provides a sub-optimal solution via iterative convex programming was proposed. As discussed in \cite{peng2012rasl}, this algorithm is limited by the following assumptions: (i) initial misalignment not too large, (ii) rank of the matrix $\bs{A}$ to be recovered not too high, and (iii) only a small fraction of all pixels affected by error. A further limitation is the scalability of the algorithm. In fact, the convex relaxation replacing the $\text{rank}(\cdot)$ with the nuclear norm requires a  very expensive Singular Value Decomposition (SVD) computation at every optimisation step.
%t-grasta
%
% \textbf{t-GRASTA.}~To address scalability issues with RASL, He et al.~\cite{he2014iterative} introduced a variant of GRASTA~\cite{he2012incremental}, referred to as t-GRASTA, which estimates the low-rank subspace, sparse error matrix, and transformation parameters such as rotation or translation. This is done by utilising iterative incremental gradient descent constrained to the Grassmann manifold of the subspaces.
% % which  to simultaneously estimate a decomposition of a collection of images into a low-rank subspace, a sparse part of occlusions and foreground objects, and a transformation such as rotation or translation of the image. 

%L2-based
% \textbf{Least-Squares Congealing (LSC).}~Despite the improvements brought by t-GRASTA in terms of scalability, methods based on low-rank minimisation for joint image alignment tend to still be relatively slow and/or dependant on a good initialisation. Least-Squares Congealing 
\textbf{Least-Squares Congealing (LSC).}~This method~\cite{cox2008least, cox2009least} has been specifically proposed for targeting large-scale joint alignment problems. Building on the success of Lucas-Kanade image alignment~\cite{lucas1981iterative}, the idea is to define a reference image $\bs{I}_j$ and align each of the remaining ones {$\{\bs{I}_i\}_{i\neq j}$} to that reference. In general, this optimisation problem can be formulated as:
\begin{equation}\label{eqn:LSC_linear}
    \argmin_{\bs{p}_{i\neq j}} \sum_{i\neq j} \Big|\Big|\bs{I}_i \circ \bs{p}_i - \bs{I}_j\Big|\Big|^2_2,
\end{equation}
where $\boldsymbol{\bs{p}} = \{\bs{p}_1, \bs{p}_2, \ldots, \bs{p}_{N-1}\}$ is the set of transformations to apply to $\{\bs{I}_i\}_{i\neq j}$ to map them onto the reference $\bs{I}_j$. The main advantage of LSC over low-rank/entropy-based ones is faster convergence, as the adoption of the least-squares cost function allows for the use of standard Gauss-Newton optimisation techniques. On the other hand, alignment performance tend to be worse due to its simplicity.  
\subsection{Proposed Method}
Motivated by the need for highly accurate alignment in very large-scale problems, we propose a congealing framework that leverages the advantages of adopting a similarity-based cost function (i.e. direct, such as $\ell_2$-norm in LSC) \textit{and} a complexity-based one (i.e. indirect, such as rank-based used in RASL). To perform this task \textit{at scale}, we formulate the congealing problem in a way that can be efficiently optimised via standard back-propagation and SGD.

Our formulation can be interpreted in terms of lossy compression optimisation~\cite{vedaldi2008joint}:
\begin{equation}\label{eqn:compression}
    \argmin_{\{\bs{p}_i\}_{i=1}^N} \mathcal{D}(\bs{I}_{i\neq j} \circ \bs{p}_{i\neq j}, \bs{I}_j) + \lambda~\mathcal{C}(\bs{I}_i \circ \bs{p}_i),
\end{equation}
where the \textit{distortion} $\mathcal{D}$ reflects the total error when approximating the reference image $\bs{I}_j$ (i.e., original data) with the aligned image $\bs{I}_i \circ \bs{p}_i$ (i.e., compressed data), the \textit{complexity} $\mathcal{C}$ is the total number of \textit{symbols} required to encode $\bs{I}_i \circ \bs{p}_i$, and the parameter $\lambda\geq0$ trades off the two quantities. 
A good candidate for our distortion (or similarity) measure $\mathcal{D}$ should be robust to occlusions, noise, outliers and,  in general, objects that might partially differ in appearance (e.g. same digit but different font, same face but wearing glasses or not). In RASL, this is achieved by adding explicit constraints on the noise model and its sparsity properties which have a significant impact on the optimisation efficiency. To circumvent this problem, we adopt the $\ell_1$-norm as measure of distortion, which can be efficiently optimised and offers a higher level of robustness compared to the $\ell_2$-norm used in LSC or \cite{vedaldi2008joint}. Formally, 
\begin{equation}\label{eqn:distortion}
    \mathcal{D} = \sum_{i\neq j}\Big|\Big|\bs{I}_{i} \circ \bs{p}_{i} - \bs{I}_j\Big|\Big|_1.
\end{equation}
Motivated by the need for optimisation in large-scale joint alignment problems, we propose an efficient alternative to rank minimisation. Specifically, we observe that when a set of images are well-aligned, they form a sequence that contains a significant level of redundant information. As a consequence, the stack of images can be compressed with higher compression rates w.r.t. the original misaligned ones. Alternatively, a lower reconstruction error can be attained, at parity of compression rate. Exploiting this consideration we therefore propose to optimise:
\begin{equation}\label{eqn:complexity}
    \begin{aligned}
      \argmin_{\{\bs{p}_i\}_{i=1}^N,\phi, \theta} &\sum_{i}\Big|\Big|D_\phi(E_\theta(\bs{I}_{i} \circ \bs{p}_{i})) - \bs{I}_{i}\circ \bs{p}_{i}\Big|\Big|_1,\\
  \textrm{s.t.} & \quad f(E_\theta(\bs{I}_{i}\circ \bs{p}_{i})) \leq \beta\\
    \end{aligned}
\end{equation}
where $\ell_1$-norm is preferred to typical $\ell_2$-norm for similar reasons as the ones mentioned above; $E_\theta:= \mathbb{R}^{w\times h} \rightarrow \mathbb{R}^b_+$ defines an encoder mapping an $w \times h$ image into a code vector $\textbf{z}$ with $b$ positive components; $D_\phi:= \mathbb{R}^b_+ \rightarrow \mathbb{R}^{w\times h}$ defines a decoder mapping a code $\textbf{z}$ into a $w \times h$ image; $f:= \mathbb{R}^b_+ \rightarrow \mathbb{R}$ defines a (monotonically increasing) positional weighting penalty applied to $\textbf{z}$. This penalty explicitly encourages the encoder-decoder to represent the aligned images using \textit{primarily} the first components of $\textbf{z}$. Similarly, $\beta$ can be interpreted as an hyper-parameter controlling the \textit{number} of first components used to represent each image, hence the representational power (or \textit{capacity}) of the encoder-decoder block. Intuitively, at parity of encoder-decoder capacity, improving the joint alignment (i.e., optimising w.r.t.~$\boldsymbol{\bs{p}}$) will lead to increased redundancy across the image stack. In fact, we would have very similar colour intensities at the same pixel location across the image stack. Therefore, this capacity will be diverted from modelling inter-image pixel intensity distributions (merely due to misalignment) to modelling the key details of the object these images share, hence leading to lower reconstruction error. With the aim of solving large-scale alignment problems efficiently, we leverage principles from Lagrangian relaxation and penalty functions~\cite{lemarechal2001lagrangian,smith1995penalty} to approximate the solution of the constrained problem in Eq.~(\ref{eqn:complexity}) and instead propose to minimise:
\begin{equation}\label{eqn:complexity2}
      \mathcal{C} = \sum_{i}\Big|\Big|D_\phi(E_\theta(\bs{I}_{i} \circ \bs{p}_{i})) - \bs{I}_{i}\circ \bs{p}_{i}\Big|\Big|_1 +~\gamma~ f(E_\theta(\bs{I}_{i}\circ \bs{p}_{i})),
\end{equation}
where $\gamma \geq 0$ trades off the contribution of the reconstruction error and the capacity of the encoder-decoder block, and $\mathcal{C}$ is our measure of complexity. Plugging Eq.~(\ref{eqn:distortion}) and Eq.~(\ref{eqn:complexity2}) in Eq.~(\ref{eqn:compression}), we obtain a novel formulation to solve congealing:
\begin{equation}\label{eqn:proposed_congealing}
    \begin{aligned}
      & \argmin_{\{\bs{p}_i\}_{i=1}^N,\phi, \theta} \sum_{i}\Big|\Big|\bs{I}_{i\neq j} \circ \bs{p}_{i\neq j} - \bs{I}_j\Big|\Big|_1 +\\
      &+ \lambda~\Big(\Big|\Big|D_\phi(E_\theta(\bs{I}_{i} \circ \bs{p}_{i})) - \bs{I}_{i}\circ \bs{p}_{i}\Big|\Big|_1 +~\gamma~f(E_\theta(\bs{I}_{i}\circ \bs{p}_{i}))\Big).
    \end{aligned}
\end{equation}
To take advantage of efficient back-propagation and SGD optimisation, (i) we implement $E_\theta$ and $D_\phi$ as Neural Networks (NNs) to form a low-capacity auto-encoder (controlled by $\gamma$); (ii) we define ${f(E_\theta(\bs{I}_{j}\circ \bs{p}_{j})) \triangleq f(\textbf{z}_j) = \textbf{w}^\top\textbf{z}_j}$, and each component $w_l$ of the weighing vector $\textbf{w} = [w_1, \ldots, w_b]^\top$ is such that $w_l = l^k / \sum_{l=1}^b l^k$ with $k \in \mathbb{N}$; and (iii) we adopt the state-of-the-art Densely fused Spatial Transformer Network (DeSTNet)~\cite{annunziata2018destnet} as the module learning and applying the set of global transformations ($\boldsymbol{\bs{p}}$) to the stack of images. Fig.~\ref{fig:flow_chart} shows the proposed method for large-scale congealing. Each input image in a batch\footnote{We use batch-based optimisation.} is first aligned to the reference $\bs{I}_j$ by the DeSTNet, and the alignment error as computed by the similarity-based loss $\mathcal{D}$ is directly back-propagated to update DeSTNet's parameters to achieve better alignment to the reference. Once a batch of images has been aligned, it goes to the penalised auto-encoder: the reconstruction error as computed by $\mathcal{C}$ is used to update (i) the auto-encoder, i.e. to improve reconstruction at parity of alignment quality, \textit{and} (ii) to further update the DeSTNet, i.e. to improve reconstruction by better alignment at parity of auto-encoder capacity. Importantly, our approach does not require gradient adjustment, as the gradient of the total loss (Eq.~(\ref{eqn:proposed_congealing})) w.r.t. the learnable parameters is implicitly and seamlessly distributed to each module
(auto-encoder and alignment), by chain-rule.
% In an attempt to leverage the advantages of the aforementioned methods, we propose to address the following optimisation problem:
% %
% \begin{equation}\label{eqn:general_congealing}
%     \begin{aligned}
%         \arg\min_{A, E, \boldsymbol{\tau}}\sum_{j\neq i}^N \Big|\Big|I_j \circ \tau_j - I_i\Big|\Big|^2_2 + \lambda~\textrm{rank}(A) \\
%         \textrm{s.t.}~~D\circ \boldsymbol{\tau} = A + E,~||E||_0 \leq k \\
%     \end{aligned}
% \end{equation}
% %
% where $\lambda$ is the hyper-parameter balancing the contribution of \textit{similarity}-based and \textit{rank}-based congealing.
% Unfortunately, this formulation would expose us to the same limitations of the RASL approach. 
%-------------------------------------------------------------------------
\section{Experiments}
We extensively evaluate the performance of the proposed method and compare it with state-of-the-art approaches~\cite{peng2012rasl, he2014iterative,oh2015partial} in terms of \textit{alignment quality}, \textit{scalability} and \textit{robustness to noise} on MNIST~\cite{lecun1998mnist} and several variants. To quantify performance, we adopt the \textit{Alignment Peak Signal to Noise Ratio}, ${\mbox{APSNR} = 10 \log_{10} \Big(\frac{255^2}{\mbox{MSE}}\Big)}$~\cite{peng2012rasl, he2014iterative,oh2015partial} where,
\begin{equation}\label{eqn:APSNR}
    \mbox{MSE} = \frac{1}{Nhw} \sum_{i=1}^{N} \sum_{r=1}^{h} \sum_{c=1}^{w}  \Big(\bs{\widehat{I}}^0_i(r,c) - \bs{\bar{I}}^0(r,c)\Big)^2,
\end{equation}
$\bs{\widehat{I}}^0_i$ represents image $i$ and $\bs{\bar{I}}^0$ the average image, both computed after alignment.  
We then investigate the impact of each individual term of the loss ($\mathcal{D}$ and $\mathcal{C}$) on the alignment quality and how they interact to achieve an improved level of performance when combined. With the aim of comparing the proposed method with \textit{Deep Congealing} (DC)~\cite{huang2012learning}\footnote{A comparison on MNIST and variants thereof was not possible as, to the best of our knowledge, the authors have not made the original implementation available.} and to assess the possibility of adopting the proposed method on more challenging datasets, we scale the framework and use it to jointly align multiple subsets of the LFW~\cite{huang2008labeled}, under different initialisation. 
\subsection{MNIST}
\label{sec:MNIST}
\begin{table*}[!tb]
\centering
\begin{small}
\caption{Architectures used for MNIST and LFW experiments. convD1-D2: convolution
layer with D1$\times$D1 receptive field, D2 channels, $\mathcal{F}$: fusion operation used in DeSTNet for fusing the parameters update, $|\bs{z}|$: dimentionality of $\bs{z}$. Default stride for convD1-D2 is $1$, $^*$ corresponds to $2$.} 
\label{tbl:architecture}
\begin{tabular}{c|c||c}
& \textbf{MNIST} & \textbf{LFW} \\ \hline
\multicolumn{1}{c|}{\textit{Aligner}} &  $\mathcal{F}$\{{[} conv$7$-$4$ $|$ conv$7$-$8$ $|$ conv$1$-$8$ {]}$\times 4\}$      &   $\mathcal{F}$\{{[} conv$3$-$64^*$ $|$ conv$3$-$128^*$ $|$ conv$3$-$256^*$ $|$ conv$1$-$8$ {]}$\times 5\}$   \\ \hline
\textit{Encoder}    & $[$conv$3$-$100^*]\times 3$ $|$ $[$conv$1$-$1024]$ $\times 2$ $|$ conv$1$-$|\bs{z}|$ &   $[$conv$3$-$128^*]\times 3$ $|$ $[$conv$1$-$512]$ $\times 2$ $|$ conv$1$-$|\bs{z}|$     \\ \hline
\textit{Decoder} &  $[$conv$1$-$1024]$ $\times 2$ $|$ conv$1$-$16$ $|$  $[$conv$3$-$100^*]\times 3$ $|$ conv$1$-$1$ &   $[$conv$1$-$512]$ $\times 2$ $|$ conv$1$-$3072$ $|$  $[$conv$3$-$128^*]\times 3$ $|$ conv$1$-$3$  \\
\end{tabular}
\end{small}
\end{table*}

\begin{figure}[!tb]
\centering
\begin{tabular}{ccccc}
 & $\scriptsize{31.74}$ & $\scriptsize{31.78}$ &
$\scriptsize{31.13}$ &
$\scriptsize{33.30}$\\
\includegraphics[align=c, width=0.072\textwidth]{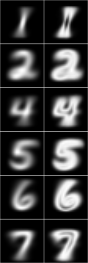} &  
\includegraphics[align=c, width=0.072\textwidth]{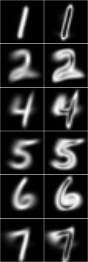}&  
\includegraphics[align=c, width=0.072\textwidth]{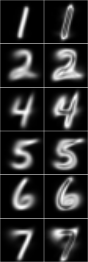}&
\includegraphics[align=c, width=0.072\textwidth]{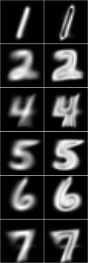}& 
\includegraphics[align=c, width=0.072\textwidth]{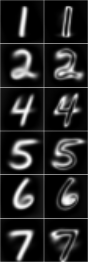} \\
(a) & (b) & (c) & (d) & (e)
\end{tabular}
\caption{Congealing results on $6\,000$ images per digit from MNIST. (a)~Before alignment, (b)~RASL~\cite{peng2012rasl}, (c)~t-GRASTA~\cite{he2014iterative}, (d)~PSSV~\cite{oh2015partial}, and (e)~Proposed method. In each subfigure (a)-(e), the first column shows means, whereas the second one shows variances. $\mbox{APSNR}$ for each digit is reported at the top of each subfigure.}
\label{fig:6k_digits_v2}
\end{figure}

\begin{figure}[!tb]
\centering
\setlength\tabcolsep{3pt}
\begin{tabular}{cccccc}
&  & $\scriptsize{31.91}$ & $\scriptsize{32.01}$ & $\scriptsize{31.40}$ & $\scriptsize{33.28}$ \\
\rotatebox[origin=c]{90}{$\sigma=10\%$}&
\includegraphics[align=c, width=0.078\textwidth]{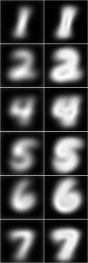} &  
\includegraphics[align=c, width=0.078\textwidth]{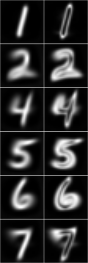}&  
\includegraphics[align=c, width=0.078\textwidth]{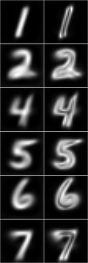}&  
\includegraphics[align=c, width=0.078\textwidth]{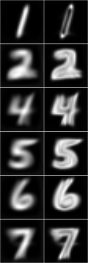}& 
\includegraphics[align=c, width=0.078\textwidth]{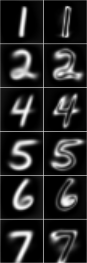}\\
&  & $\scriptsize{31.66}$ & $\scriptsize{30.64}$ & $\scriptsize{30.48}$ & $\scriptsize{33.20}$\\
\rotatebox[origin=c]{90}{$\sigma=20\%$}&
\includegraphics[align=c, width=0.078\textwidth]{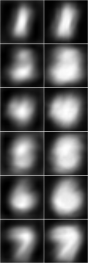} &  
\includegraphics[align=c, width=0.078\textwidth]{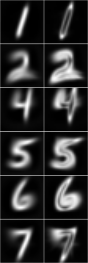}&  
\includegraphics[align=c, width=0.078\textwidth]{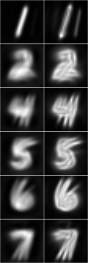}&  
\includegraphics[align=c, width=0.078\textwidth]{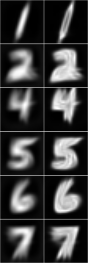}& 
\includegraphics[align=c, width=0.078\textwidth]{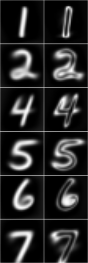}\\
&  & $\scriptsize{29.80}$ & $\scriptsize{29.36}$ & $\scriptsize{29.47}$ & $\scriptsize{32.96}$\\
\rotatebox[origin=c]{90}{$\sigma=30\%$}&
\includegraphics[align=c, width=0.078\textwidth]{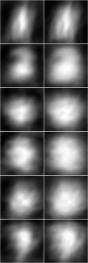} &  
\includegraphics[align=c, width=0.078\textwidth]{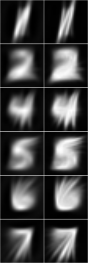}&  
\includegraphics[align=c, width=0.078\textwidth]{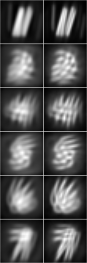}&  
\includegraphics[align=c, width=0.078\textwidth]{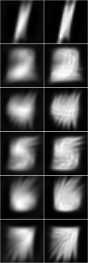}&  
\includegraphics[align=c, width=0.078\textwidth]{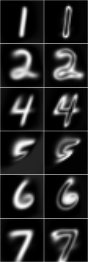}\\
& (a) & (b) & (c) & (d) & (e)
\end{tabular}
\caption{Robustness of congealing methods to random perspective warps with $\sigma = \big\{10\%, 20\%, 30\%\big\}$, corresponding to top, middle and bottom block, respectively. (a)~Before alignment, (b)~RASL~\cite{peng2012rasl}, (c)~t-GRASTA~\cite{he2014iterative}, (d)~PSSV~\cite{oh2015partial}, and (e)~Proposed method. In each subfigure (a)-(e), the first column shows means, whereas the second one shows variances. For compactness, $\mbox{APSNR}$ for each method is averaged across the digits and reported at the top of each cell.}
\label{fig:mnist_pertr}
\end{figure}

With the aim of evaluating the scalability of the proposed method and the baselines, we start by creating multiple MNIST subsets, as follows. For each digit in $\{0, 1, 2, 3, 4, 5, 6, 7, 8, 9\}$, we randomly sample $\{1\,000, 2\,000, 3\,000, 4\,000, 5\,000, 6\,000\}$ images from the original MNIST dataset and align them separately. For the proposed method, we adopt \textit{DeSTNet-4}~\cite{annunziata2018destnet} with expansion rate $k^F=32$ as aligner, and the penalised reconstruction auto-encoder defined in Table~\ref{tbl:architecture}, where we use \textit{tanh} non-linearities after each layer, apart from the last layer of the encoder, where \textit{sigmoid} is used, to keep each component of \textbf{z} in $\big[0, 1\big]$. We set $\lambda=1$ to use both similarity- and complexity-based loss, $\gamma=1$ and $k=1$. We optimise the entire architecture end-to-end, using a standard Adam-based SGD optimiser with learning rate $10^{-5}$.
\begin{figure}[!tb]
\centering
\setlength\tabcolsep{1.5pt}
\begin{tabular}{c}
\includegraphics[align=c, width=0.45\textwidth]{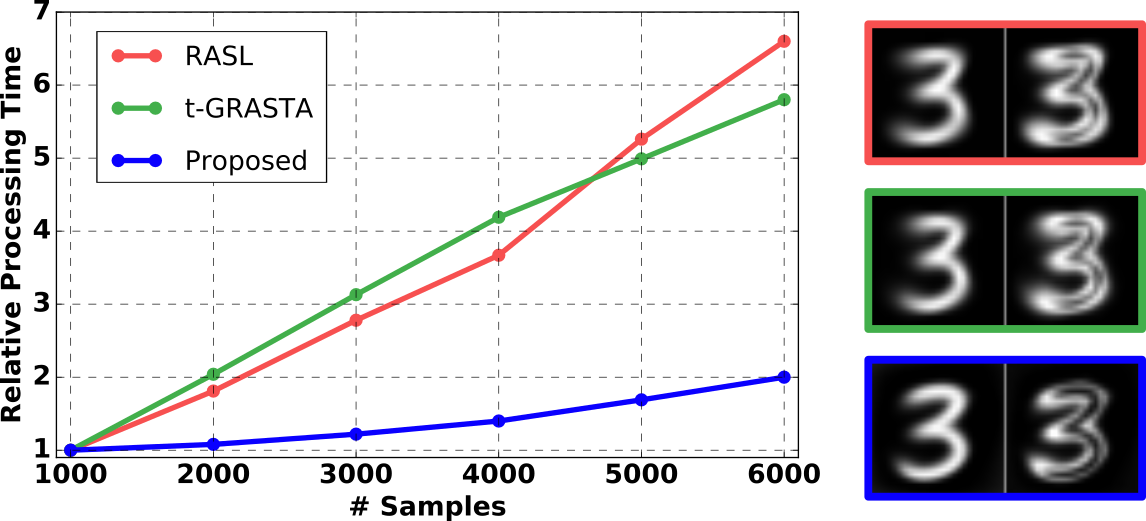}
\end{tabular}
\caption{Relative processing time for RASL~\cite{peng2012rasl}, t-GRASTA~\cite{he2014iterative}, and the proposed method when aligning an increasingly large number of images. Mean and variances of the aligned images produced by the compared methods for the $6\,000$ samples are also displayed.}
\label{fig:processing_time}
\end{figure}
Following~\cite{peng2012rasl, he2014iterative, learned2006data}, we qualitatively assess alignment results for the proposed method and the baselines by computing the mean and variance across the entire dataset before and after alignment. To evaluate \textit{scalability}, we measure the relative processing time for RASL, t-GRASTA, and the proposed method when aligning an increasingly large number of images. Due to the difference in hardware (CPUs used by the baselines, GPUs by the proposed method), we normalise processing times w.r.t. the time required to align $1,000$ images to provide a fair comparison. As Fig.~\ref{fig:processing_time} shows for the case of digit `$3$'\footnote{Similar results hold for the other digits.}, the proposed method scales better than the baselines. Moreover, as Fig.~\ref{fig:6k_digits_v2} shows in the most challenging case, i.e. datasets with $6,000$ images, the much sharper mean and lower variance images (hence higher \mbox{APSNR}) suggest that proposed method achieves much better alignment too.
Following the experimental protocol in~\cite{lin2017inverse, annunziata2018destnet}, we evaluate the robustness of each method to synthetic distortions based on random perspective warps. Specifically, assuming each MNIST image is $s\times s$ pixels ($s = 28$), the four corners of each image are independently and randomly scaled with Gaussian noise $\mathcal{N}(0, \sigma^2s^2)$, then randomly translated with the same noise model. We assess alignment quality under three levels of perturbation, i.e. $\sigma = \big\{10\%, 20\%, 30\% \big\}$. To this aim, we apply this perturbation model to each  $6\,000$ images dataset and report a subset of the results in 
Fig.~\ref{fig:mnist_pertr}.
We observe that although a $10\%$ perturbation seems to be well handled by RASL and {t-GRASTA}, alignment performance deteriorates significantly at $20\%$ and they tend to fail at the most challenging $30\%$. On the other hand, the proposed method shows strong robustness to this perturbation model across all the digits and under significant noise.
%
% Those results suggest that, while a $10\%$ perturbation seem to be well handled by \textit{RASL} and \textit{t-GRASTA}, alignment performance deteriorate significantly at $20\%$ and they completely fail at the most challenging $30\%$. On the other hand, the proposed method shows strong robustness to this perturbation model across all the digits and under significant noise.
%
\subsection{Ablation Study}
\begin{figure}[!tb]
\centering
\setlength\tabcolsep{1.5pt}
\begin{tabular}{cccc}
\includegraphics[align=c, width=0.08\textwidth]{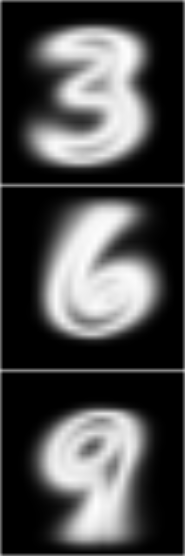} & 
\includegraphics[align=c, width=0.08\textwidth]{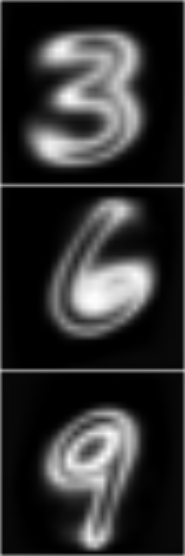} & 
\includegraphics[align=c, width=0.08\textwidth]{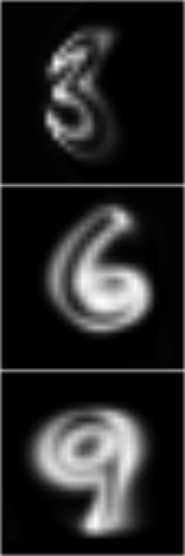} & 
\includegraphics[align=c, width=0.08\textwidth]{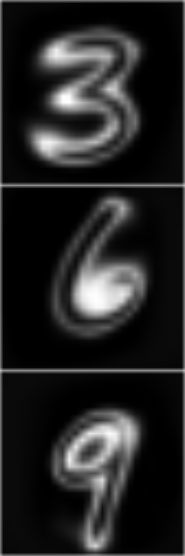}\\
(a) & (b) & (c) & (d)
\end{tabular}
\caption{Ablation study: disentangling the impact of the similarity- ($\mathcal{D}$) and complexity-based ($\mathcal{C}$) losses on the final alignment result. Variance images (a) before alignment, (b) $\mathcal{D}$-only, (c) $\mathcal{C}$-only, and (d) both.}
\label{fig:losses}
\end{figure}

The proposed congealing approach takes advantage of both the similarity- and complexity-based losses (i.e., $\mathcal{D}$ and $\mathcal{C}$ in Eq.~(\ref{eqn:distortion}) and Eq.~(\ref{eqn:complexity2}), respectively), as described in Eq.~(\ref{eqn:proposed_congealing}). With the aim of disentangling the contribution of each term to the final result, we have evaluated the joint alignment performance when one of the two losses is excluded from the optimisation. Figs.~\ref{fig:losses}(b) and (c) show the alignment results when excluding $\mathcal{D}$, and $\mathcal{C}$, respectively, while the alignment results produced when both are used are displayed in Fig.~\ref{fig:losses}(d). We observe that, in general, excluding $\mathcal{D}$ has a stronger impact on the final alignment results; moreover, the use of the reference image when computing $\mathcal{D}$ makes the optimisation much more robust, as it implicitly avoids the shrinking effect typically observed when only $\mathcal{C}$ is used. The latter is due to the fact that, at parity of reconstruction capacity for the auto-encoder, a lower complexity measure is attained when the object to reconstruct shows less spatial variability and can therefore be better reconstructed\footnote{Notice that this undesired effect is typical of low-rank-based congealing approaches~\cite{peng2012rasl}.} (see Eq.~(\ref{eqn:complexity2})). 
We observe that, (i) the addition of $\mathcal{C}$ to the loss based only on $\mathcal{D}$, contributes to further refining the alignment results and achieving even lower variance (see digit `$6$' and `$9$'); (ii) importantly, $\mathcal{C}$ tends to drive the overall optimisation towards solutions that favour a more (spatially) uniform alignment, as shown for digit `$3$'; in this sense, the complexity-based loss can be interpreted as a regulariser.
\subsection{affNIST}
\begin{figure}[!tb]
\centering
\setlength\tabcolsep{2pt}
\begin{tabular}{cccc}
$\scriptsize{33.19}$ & $\scriptsize{33.29}$ & $\scriptsize{34.44}$ & $\scriptsize{32.59}$ \\
\includegraphics[align=c, width=0.1\textwidth]{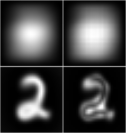} &  \includegraphics[align=c, width=0.1\textwidth]{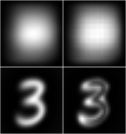}&  \includegraphics[align=c, width=0.1\textwidth]{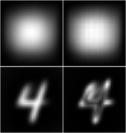}&  \includegraphics[align=c, width=0.1\textwidth]{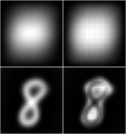}\vspace{-0.3cm}
\\ \\
\includegraphics[align=c, width=0.1\textwidth]{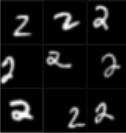} &  \includegraphics[align=c, width=0.1\textwidth]{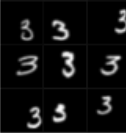}&  \includegraphics[align=c, width=0.1\textwidth]{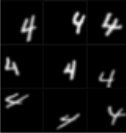}&  \includegraphics[align=c, width=0.1\textwidth]{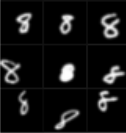}\vspace{-0.3cm}
\\ \\
\includegraphics[align=c, width=0.1\textwidth]{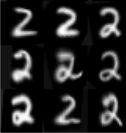} &  \includegraphics[align=c, width=0.1\textwidth]{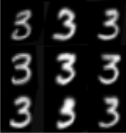}&  \includegraphics[align=c, width=0.1\textwidth]{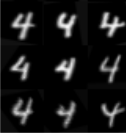}&  \includegraphics[align=c, width=0.1\textwidth]{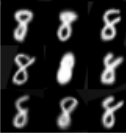}\\
(a) & (b) & (c) & (d)
\end{tabular}
\caption{Congealing results of the proposed method on $100\,000$ images per digit from affNIST. (a)-(d) correspond to different digits. Top: mean (first columns) and variance (second columns) images, before (first rows) and after (second rows) alignment. Middle: a subset of the actual inputs. Bottom: a subset of the actual outputs. $\mbox{APSNR}$ for each digit is reported at the top of each subfigure.}
\label{fig:affnist}
\end{figure}

Previously proposed congealing approaches have shown limitations in terms of scaling efficiency; in fact, on very low-resolution datasets, joint alignment optimisation results have been reported only for up to a few thousands samples~\cite{cox2009least}. Moreover, as confirmed in the experiments reported in the previous section, large intra-class spatial variability (modelled with synthetic perturbation) seems to significantly deteriorate performance. To further push the limits and evaluate the performance of the proposed method, we assess joint alignment performance on a much more challenging version of MNIST, namely $\textit{affNIST}$~\cite{afnist}. This dataset is built by taking images from MNIST and applying various reasonable affine transformations to them. In the process, the images become $40 \times 40$ pixels large, with significant translations involved. From this dataset, we take the first $100\,000$ samples for each digit and perform alignment (results in Fig.~\ref{fig:affnist}), using the same parameter setting adopted in the experiments above. The strong variability characterising this dataset is clear by looking at the means and variances before alignment, and a subset of the actual inputs (Fig.~\ref{fig:affnist}-middle). Nevertheless, the proposed method achieves a good level of alignment, as demonstrated by the average and variance images after alignment (hence high \mbox{APSNR}) and a subset of the actual outputs (Fig.~\ref{fig:affnist}-bottom).
%
% \begin{figure}[h]
% \centering
% \setlength\tabcolsep{2pt}
% \begin{tabular}{ccccc}
% \includegraphics[align=c, width=0.1\textwidth]{figures/figure_4/2_o.png} &  \includegraphics[align=c, width=0.1\textwidth]{figures/figure_4/2_i.png}&  \includegraphics[align=c, width=0.1\textwidth]{figures/figure_4/2_a.png}&  \includegraphics[align=c, width=0.1\textwidth]{figures/figure_4/2_r.png}\\
% \includegraphics[align=c, width=0.1\textwidth]{figures/figure_4/3_o.png} &  \includegraphics[align=c, width=0.1\textwidth]{figures/figure_4/3_i.png}&  \includegraphics[align=c, width=0.1\textwidth]{figures/figure_4/3_a.png}&  \includegraphics[align=c, width=0.1\textwidth]{figures/figure_4/3_r.png}\\
% \includegraphics[align=c, width=0.1\textwidth]{figures/figure_4/4_o.png} &  \includegraphics[align=c, width=0.1\textwidth]{figures/figure_4/4_i.png}&  \includegraphics[align=c, width=0.1\textwidth]{figures/figure_4/4_a.png}&  \includegraphics[align=c, width=0.1\textwidth]{figures/figure_4/4_r.png}\\
% \includegraphics[align=c, width=0.1\textwidth]{figures/figure_4/8_o.png} &  \includegraphics[align=c, width=0.1\textwidth]{figures/figure_4/8_i.png}&  \includegraphics[align=c, width=0.1\textwidth]{figures/figure_4/8_a.png}&  \includegraphics[align=c, width=0.1\textwidth]{figures/figure_4/8_r.png}\\
% (a) & (b) & (c) & (d)
% \end{tabular}
% \caption{affnist}
% \label{fig:affnist}
% \end{figure}
%
\subsection{infiMNIST}
\begin{figure}[!tb]
\centering
\setlength\tabcolsep{1.5pt}
\begin{tabular}{ccc}
$\scriptsize{32.20}$ & $\scriptsize{32.63}$ & $\scriptsize{31.93}$ \\
\includegraphics[align=c, width=0.1\textwidth]{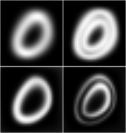}&  
\includegraphics[align=c, width=0.1\textwidth]{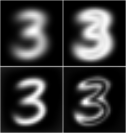}& 
\includegraphics[align=c, width=0.1\textwidth]{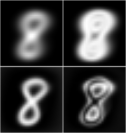}\vspace{-0.3cm}
\\ \\
\includegraphics[align=c, width=0.1\textwidth]{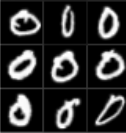} &  
\includegraphics[align=c, width=0.1\textwidth]{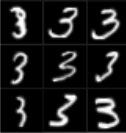}&  
\includegraphics[align=c, width=0.1\textwidth]{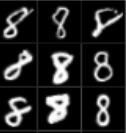}\vspace{-0.3cm}
\\ \\
\includegraphics[align=c, width=0.1\textwidth]{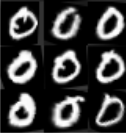}&
\includegraphics[align=c, width=0.1\textwidth]{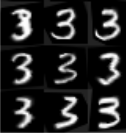}&  
\includegraphics[align=c, width=0.1\textwidth]{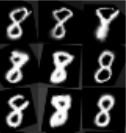}\\
(a) & (b) & (c)
\end{tabular}
\caption{Congealing results of the proposed method on $1\,000\,000$ images per digit from \mbox{infiMNIST}. (a)-(c) correspond to different digits. Top: mean (first columns) and variance (second columns) images, before (first rows) and after (second rows) alignment. Middle: a subset of the actual inputs. Bottom: a subset of the actual outputs. $\mbox{APSNR}$ for each digit is reported at the top of each subfigure.}
\label{fig:infnist}
\end{figure}

So far, the proposed method has shown robustness to global affine/perspective perturbations, and joint alignment problems with up to $100,000$ samples per digit. Here, we evaluate the alignment performance under non-linear (local) deformations (e.g. tickening) and translations, and solve the joint alignment problem for $1,000,000$ images per digit sampled from \mbox{infiMNIST}~\cite{loosli2007training}\footnote{The code to generate datasets from \mbox{infiMNIST} is available at \href{https://leon.bottou.org/projects/\mbox{infiMNIST}}{https://leon.bottou.org/projects/\mbox{infiMNIST}}.}. Notice, we use the same parameter setting adopted above to assess the robustness and generalisation of the proposed method in a much more challenging joint alignment problem. As Fig.~\ref{fig:infnist} shows, despite the random translations being relatively smaller than the ones used in affNIST, the non-linear perturbations add a much higher level of intra-class variability. Nevertheless, the proposed method achieves remarkable joint alignment at this scale and under this kind of perturbations. 
%
% \begin{figure}[h]
% \centering
% \setlength\tabcolsep{1.5pt}
% \begin{tabular}{cccc}
% \includegraphics[align=c, width=0.09\textwidth]{figures/figure_6/all_digits_overall.png} & 
% \includegraphics[align=c, width=0.09\textwidth]{figures/figure_6/all_digits_input.png} & 
% \includegraphics[align=c, width=0.09\textwidth]{figures/figure_6/all_digits_aligned.png} & 
% \includegraphics[align=c, width=0.09\textwidth]{figures/figure_6/all_digits_reconstructed.png} \\
% (a) & (b) & (c) & (d)
% \end{tabular}
% \caption{infnist}
% \label{fig:infnist}
% \end{figure}
%
\subsection{LFW}
\begin{figure}[!tb]
\centering
\setlength\tabcolsep{1.5pt}
\begin{tabular}{ccccc}
& $\scriptsize{29.90}$ & $\scriptsize{29.74}$ & $\scriptsize{30.31}$ & $\scriptsize{30.73}$\\
\includegraphics[align=c, width=0.09\textwidth]{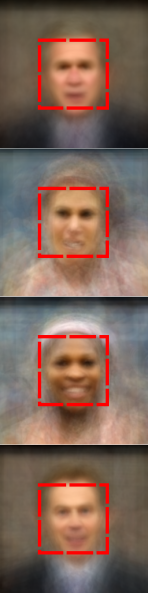} & 
\includegraphics[align=c, width=0.09\textwidth]{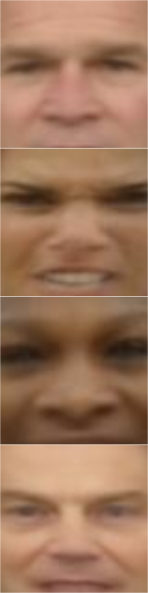} & 
\includegraphics[align=c, width=0.09\textwidth]{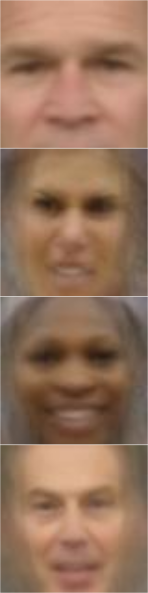} & 
\includegraphics[align=c, width=0.09\textwidth]{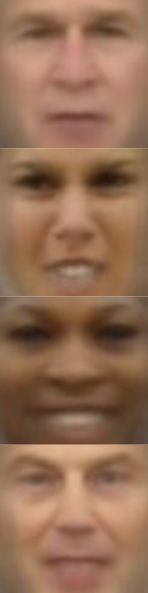} & 
\includegraphics[align=c, width=0.09\textwidth]{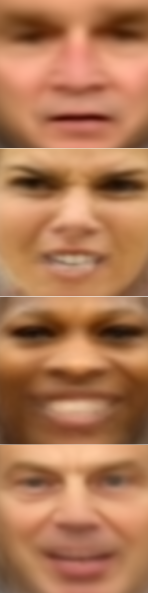} \\
(a) & (b) & (c) & (d) & (e)
\end{tabular}
\caption{Congealing results (means) on LFW. (a) Before alignment, (b) RASL~\cite{peng2012rasl}, (c)~PSSV~\cite{oh2015partial}, (d) Deep Congealing~\cite{huang2012learning}, and (e) Proposed method. The bounding box initialisation is shown in red in (a) for all the subjects. For compactness, average $\mbox{APSNR}$ for each method is reported at the top of each subfigure and averaged across the subjects.}
\label{fig:lfw_rasl_nips_ours}
\end{figure}

LFW~\cite{learned2016labeled} has been widely used to assess the performance of state-of-the-art joint alignment methods, e.g. in~\cite{peng2012rasl,huang2012learning}. This dataset is made challenging by multiple factors, including variations in facial expression, occlusion, illumination changes, clutter in the background and head pose variations. Moreover, each subject image is $250 \times 250$ pixels, which is much larger than MNIST (and variants) images used in the experiments above. We selected four subsets, corresponding to male and female subjects with the largest amount of images, namely \texttt{George\_W\_Bush}, \texttt{Tony\_Blair}, \texttt{Serena\_Williams}, and \texttt{Jennifer\_Capriati}. To accommodate the difference in input image size and considering the more complex task w.r.t. MNIST-based datasets, we scale the aligner and the encoder-decoder block as reported in Table~\ref{tbl:architecture}.
In Fig.~\ref{fig:lfw_rasl_nips_ours}, we report a qualitative and quantitative comparison of the proposed method with RASL~\cite{peng2012rasl}, PSSV~\cite{oh2015partial} and Deep Congealing~\cite{huang2012learning}, for which joint alignment results initialised with the Viola-Jones face detector~\cite{viola2004robust} are available at \href{http://vis-www.cs.umass.edu/lfw/}{http://vis-www.cs.umass.edu/lfw/}. For fair comparison, we adopt the same initialisation for the proposed method and the baselines. We observe that, overall, the proposed method outperforms both RASL, PSSV and Deep Congealing, in terms of $\mbox{APSNR}$ which is qualitatively confirmed by sharper average images across all the subjects. Moreover, unlike RASL and PSSV, the proposed method does not suffer a zoom-in/zoom-out effect which makes the optimisation focus on smaller/larger portion of the region of interest. This can be attributed to the use of the reference image in $\mathcal{D}$. 

\begin{figure}[!tb]
\centering
\setlength\tabcolsep{1.5pt}
\begin{tabular}{ccccccc}
Init~\cite{viola2004robust} & \multicolumn{3}{c}{$\longleftarrow+15\%\longrightarrow$} & \multicolumn{3}{c}{$\longleftarrow+30\%\longrightarrow$}\\
& $\scriptsize{29.75}$ & $\scriptsize{29.25}$ & $\scriptsize{30.43}$ & $\scriptsize{29.14}$ & $\scriptsize{28.67}$ & $\scriptsize{30.29}$\\
\includegraphics[align=c, width=0.06\textwidth]{figures/figure_8_b/input.png} & 
\includegraphics[align=c, width=0.06\textwidth]{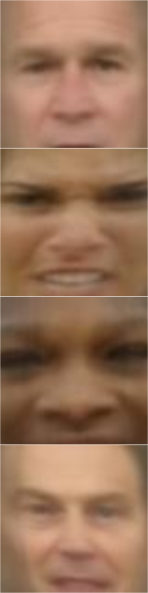} & 
\includegraphics[align=c, width=0.06\textwidth]{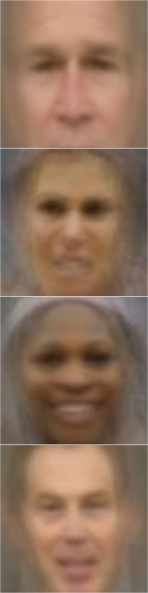} & 
\includegraphics[align=c, width=0.06\textwidth]{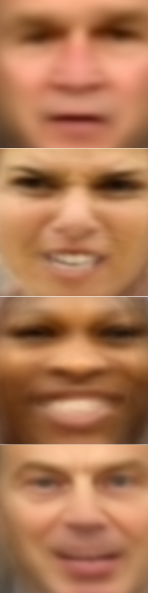} & 
\includegraphics[align=c, width=0.06\textwidth]{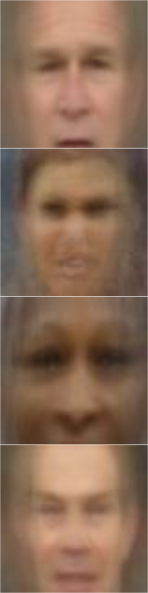} & 
\includegraphics[align=c, width=0.06\textwidth]{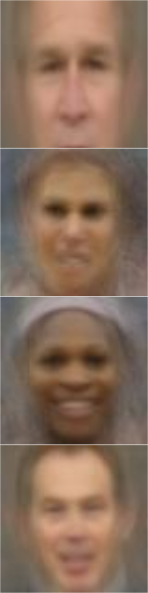} & 
\includegraphics[align=c, width=0.06\textwidth]{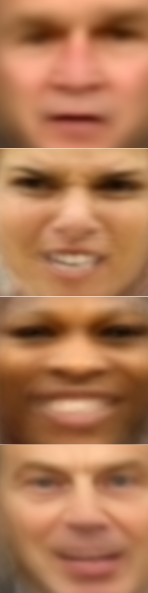}\\
(a) & (b) & (c) & (d) & (e) & (f) & (g)
\end{tabular}
\caption{Robustness of congealing methods to initialisation, i.e. bounding box $15\%$ and $30\%$ larger than the one estimated by~\cite{viola2004robust} (in red colour in (a)). Mean images (a) before alignment, (b)(e) RASL~\cite{peng2012rasl}, (c)(f)~PSSV~\cite{oh2015partial}, and (d)(g)~Proposed method. For compactness, average $\mbox{APSNR}$ for each method is reported at the top of each subfigure and averaged across the subjects.}
\label{fig:rasl_our_m_l}
\end{figure}

Although important progress has been made in recent years in face detection~\cite{chen2016supervised, zafeiriou2015survey, zhang2016joint,sun2018face}, some level of inaccuracy is inevitable in a practical setting. So, it is important to assess the robustness of the proposed method to coarser initialisation. To this aim, we increased the size of the initial bounding box returned by the Viola-Jones face detector by $15\%$ and $30\%$ in width and height, and report the joint alignment results in Fig.~\ref{fig:rasl_our_m_l}. We observe that the performance of both RASL (Figs~\ref{fig:rasl_our_m_l}(b,e)) and PSSV (Figs~\ref{fig:rasl_our_m_l}(c,f)) degrade significantly when the initialisation is not close to the object, as confirmed by a sharper decrease in average $\mbox{APSNR}$ and the average aligned faces being blurry. Instead, the proposed method demonstrates strong robustness to the initialisation: as can be observed in Figs~\ref{fig:rasl_our_m_l}(d,g), our mean aligned faces are clean and crisp which indicates a remarkable level of alignment even with a bounding box $30\%$ larger. 

Following the protocol adopted in \cite{peng2012rasl, he2014iterative}, we further quantify alignment performance by computing the average errors in the locations of three landmarks (the eye outer corners and tip of nose), calculated as the distances of the estimated locations to their centre, normalised by the eye-to-eye distance. We compare our alignment performance against RASL (best rank-based baseline) and DC (deep learning approach). We average the performance for each landmark in a given subject and report them in Table~\ref{tbl:landmarks_lfw}. Confirming the considerations above, when the original initialisation is adopted, the proposed method attains the lowest errors across all the subjects. Moreover, while at $15\%$ coarser initialisation RASL starts to show difficulties on some subjects, at $30\%$ performance degrades significantly. Instead, the proposed method shows much stronger robustness across subjects and initialisation.
\begin{table}[!tb]
\centering
\begin{small}
\caption{Average errors for three landmarks (the eye outer corners and tip of nose), calculated as the distances of the estimated locations to their centre, normalised by the eye-to-eye distance. S1:\texttt{George\_W\_Bush}, S2:\texttt{Jennifer\_Capriati}, 
S3:\texttt{Serena\_Williams}, S4:\texttt{Tony\_Blair}.}
\label{tbl:landmarks_lfw}
\begin{tabular}{c|c|c|c|c|c}
     \textbf{Init}              &    \textbf{Methods}      & \textbf{S1}              & \textbf{S2}              & \textbf{S3}              & \textbf{S4}              \\ \hline
\multirow{4}{*}{\cite{viola2004robust}} & RASL~\cite{peng2012rasl}     & 2.88\%          & 2.45\%          & 3.32\%          & 3.24\%          \\ 
                   & DC~\cite{huang2012learning}     & 3.97\%          & 3.48\%          & 3.48\%          & 3.27\%          \\ 
                   & Proposed & \textbf{2.67\%} & \textbf{1.86\%} & \textbf{2.24\%} & \textbf{2.39\%} \\ \hline \hline
\multirow{3}{*}{\textbf{$+15\%$}} & RASL~\cite{peng2012rasl}     & \textbf{3.24\%} & 6.40\%          & 5.02\%          & 3.65\%          \\ 
                   & Proposed & 3.84\%          & \textbf{2.12\%} & \textbf{4.34\%} & \textbf{2.04\%} \\ \hline \hline
\multirow{3}{*}{\textbf{$+30\%$}} & RASL~\cite{peng2012rasl}     & 6.29\%          & 6.77\%          & 7.08\%          & 6.87\%          \\  
                   & Proposed & \textbf{4.27\%} & \textbf{1.92\%} & \textbf{3.69\%} & \textbf{2.55\%} \\
\end{tabular}
\end{small}
\end{table}
%
%---------------------------------------------------------------------------------
\section{Conclusions}
Image alignment is a major area of research in computer vision. However, the majority of previously proposed methods focus on identifying pixel-level correspondences between a \textit{pair} of images. Instead, a plethora of other tasks such as, co-segmentation, image edit propagation and structure-from-motion, would considerably benefit from establishing pixel-level correspondences between a \textit{set} of images. Several congealing or joint alignment methods have been previously proposed; however, scalability to large datasets and the limited robustness to initialisation and intra-class variability seem to hamper their wide applicability. To address these limitations, we have proposed a novel congealing method and shown that it is capable of handling joint alignment problems at very large scale i.e., up to one million data points, simultaneously. This is achieved through a novel \textit{differentiable} formulation of the congealing problem, which combines the advantages of similarity- and rank-based congealing approaches and can be easily optimised with standard SGD, end-to-end. Extensive experimental results on several benchmark datasets, including digits and faces at different resolutions, show that the proposed congealing framework outperforms state-of-the-art approaches in terms of scalability, alignment quality and robustness to linear and non-linear geometric perturbations of different magnitude and type.   
%-------------------------------------------------------------------------
{\small
\bibliographystyle{ieee_fullname}
\bibliography{egbib}
}
\end{document}